\documentclass[runningheads]{llncs}
\usepackage{graphicx}
\usepackage{amsmath,amssymb} 
\usepackage{nccmath}
\usepackage{color}
\usepackage[font={footnotesize}]{caption}
\usepackage{subcaption}
\usepackage{url}
\usepackage[width=122mm,left=12mm,paperwidth=146mm,height=193mm,top=12mm,paperheight=217mm]{geometry}
\usepackage{pdfpages}
\begin{document}
\pagestyle{headings}
\mainmatter

\title{Collaborative Layer-wise Discriminative Learning in Deep Neural Networks} 

\titlerunning{Collaborative Layer-wise Discriminative Learning in Deep Neural Networks}

\authorrunning{Xiaojie Jin, Yunpeng Chen, Jian Dong, Jiashi Feng, Shuicheng Yan}

\author{Xiaojie Jin$^1$ Yunpeng Chen$^2$ Jian Dong$^3$ Jiashi Feng$^2$ Shuicheng Yan$^{3,2}$}


\institute{\small $^1$NUS Graduate School for Integrative Science and Engineering, NUS\\
\small$^2$Department of ECE, NUS \qquad $^3$360 AI Institute\\
\tt\small \{xiaojie.jin, chenyunpeng, elefjia\}@nus.edu.sg\\
\tt\small \{dongjian-iri, yanshuicheng\}@360.cn}

\maketitle
\vspace{-5mm}

\begin{abstract} Intermediate features at different layers of a deep
  neural network are known to be discriminative for visual patterns of
  different complexities. However, most existing works ignore such
  cross-layer heterogeneities when classifying samples of different
  complexities. For example, if a training sample has already been
  correctly classified at a specific layer with high confidence, we
  argue that it is unnecessary to enforce rest layers to classify this
  sample correctly and a better strategy is to encourage those layers
  to focus on other samples.

  \ \ \quad In this paper, we propose a layer-wise discriminative
  learning method to enhance the discriminative capability of a deep
  network by allowing its layers to work collaboratively for
  classification. Towards this target, we introduce multiple
  classifiers on top of multiple layers. Each classifier not only
  tries to correctly classify the features from its input layer, but
  also coordinates with other classifiers to jointly maximize the
  final classification performance. Guided by the other companion
  classifiers, each classifier learns to concentrate on certain
  training examples and boosts the overall performance. Allowing for
  end-to-end training, our method can be conveniently embedded into
  state-of-the-art deep networks. Experiments with multiple popular
  deep networks, including Network in Network, GoogLeNet and VGGNet,
  on scale-various object classification benchmarks, including
  CIFAR100, MNIST and ImageNet, and scene classification benchmarks,
  including MIT67, SUN397 and Places205, demonstrate the effectiveness
  of our method. In addition, we also analyze the relationship between
  the proposed method and classical conditional random fields models.
\end{abstract}
\vspace{-8.5mm}
\section{Introduction}
\vspace{-0.5mm} In recent years, deep neural networks (DNNs) have
achieved great success in a variety of machine learning
tasks~\cite{krizhevsky2012imagenet,RCNN,fullyconvseg,residual,zhou2014learning,cirecsan2011convolutional,karpathy2015deep,guosen,hcp}. One
of the critical advantages contributing to the spectacular
achievements of DNNs is their strong capability to automatically learn
hierarchical feature representations from a large amount of training
data~\cite{bengioftml,hintonscience,farabethier,leeconvhier}, which
hence allows the deep models to build sophisticated and highly
discriminative features without the harassment of hand-feature
engineering. It is well known that deep models learn increasingly
abstract and complex concepts from the bottom input layer to the top
output layer~\cite{visualhier,bengiobook}. Generally, deep models
learn low-level features in bottom layers, such as corners, lines and
circles, then mid-level features such as textures and object parts in
intermediate layers, and finally semantically meaningful concepts in
top layers, e.g. the spatial geometry in a scene
image~\cite{zhou2014learning} and the structure of an object, e.g. a
face \cite{closefacegap}. In other words, the features learned by deep
models, being discriminative for visual patterns of different
complexities, are distributed across the whole network.

\vspace{-0.5mm}

\begin{figure}[!t] \centering
\centering
	\includegraphics[width=12cm]{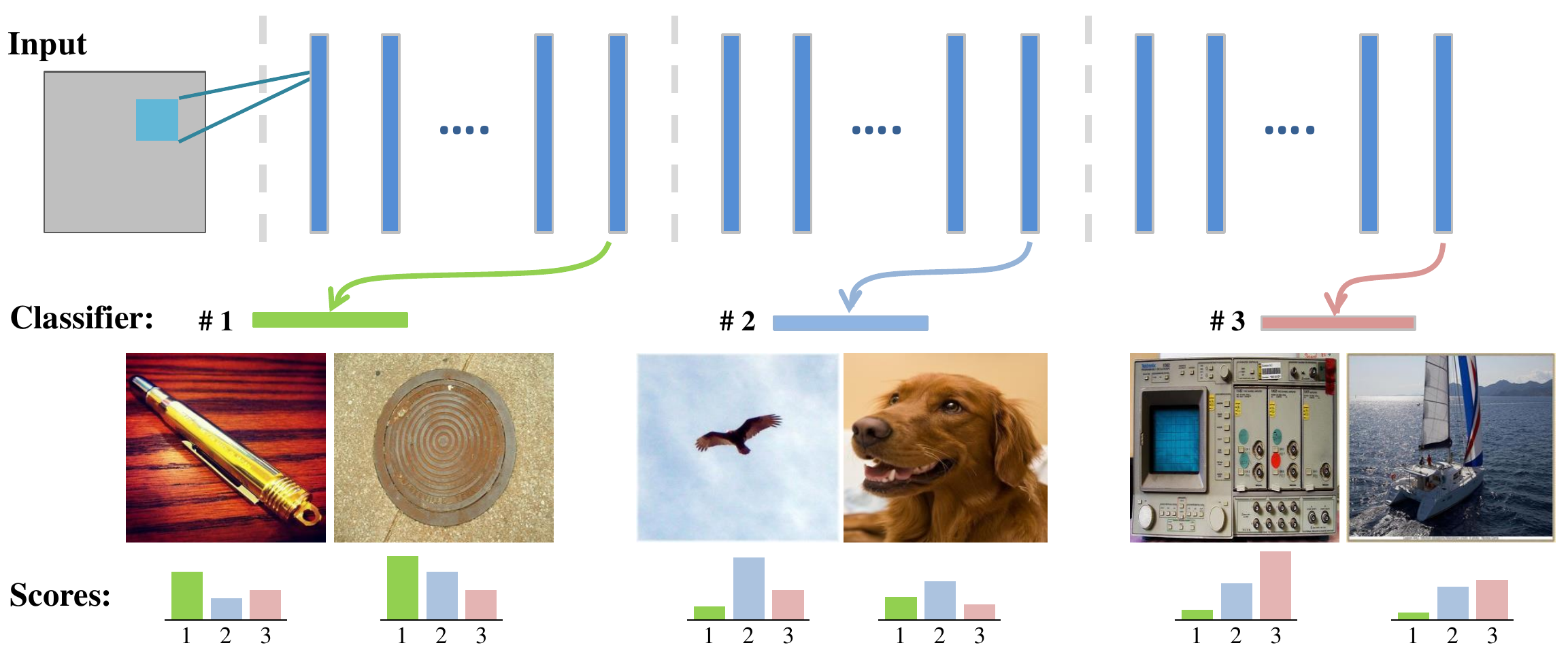}
	\vspace{-1mm}
	\caption{Examples from ImageNet dataset showing that features
          from different layers in a deep network are good at
          discriminating images of different complexities.  Three
          collaborative classifiers are introduced at different layers
          of a deep model using our proposed CLDL method. The input
          images in the middle row are with increasing complexity from
          left to right. The bottom row shows the prediction scores of
          corresponding images on the ground truth category produced
          by classifiers in CLDL. One can find that classifiers
          introduced at bottom/top layers of a deep model can
          correctly classify simple/complex samples. Note that
          classifiers with a smaller index number lie at lower-level
          layers. All figures in this paper are best viewed in color.}
	\vspace{-5.6mm}
	\label{fig:featurevis}
\end{figure}

However, although such a hierarchical property of learned features by
deep models has been recognized for a long time, most of existing
works \cite{krizhevsky2012imagenet,zhou2014learning,vgg} only use
features from the top output layer and ignore such heterogeneity
across different layers. We propose a better policy based on the
following consideration: in the task of classifying multiple
categories, for many simple input samples, the features represented in
bottom or intermediate layers already have sufficient discriminative
capability for classification. For example, in the fine-grained
classification task, correctly recognizing objects with small
intra-class variance like bird species and flower species largely
depends on fine-scale and local input features like the color
difference and shape distortion, which are easily ignored by top
layers because they tend to learn semantic features. Another example
is scene classification, where features in the intermediate layer may
be sufficiently good for classifying object-centric scene categories,
e.g. discriminating a bedroom from other scenes through extracting
features around a bed. The top output layer may be inclined to learn
the spatial configuration of scenes. Fig.~\ref{fig:featurevis}
provides more examples. Some recently published works also provide
similar observations. Yang et al.~\cite{dagcnn} showed that different
categories of scene images are best classified by features extracted
from different layers. In~\cite{hypercolumns}, it has been verified
that considering mid-level or low-level features increases the
segmentation and localization accuracy. However, those works just take
features from different layers together and feed the combined features
into a single classifier. This strategy may impede the further
performance improvement as verified in our experiments due to the
introduced redundant information from less discriminative layers.

\vspace{-0.5mm}

In this paper, aiming to fully utilize the knowledge distributed in
the whole model and boost the discriminative capability of deep
networks, we propose a {\bf C}ollaborative {\bf L}ayer-wise {\bf
  D}iscriminative {\bf L}earning (CLDL) method that allows classifiers
at different layers to work jointly in the classification task. The
resulted model trained by CLDL is called CLDL-DNN. Our method is
motivated by the following rationale: in training a deep network
model, if a sample has already been correctly classified at a specific
layer with high confidence, it is unnecessary to enforce the rest
layers to focus on classifying this sample correctly and we propose to
let them focus on other samples that are not classified correctly
yet. More concretely, to implement this idea, we introduce multiple
classifiers on top of multiple layers. Each classifier not only tries
to correctly classify the features from its input layer, but also
coordinates with other classifiers to jointly maximize the final
classification performance. Guided by the other companion classifiers,
each classifier learns to concentrate on certain training
examples. Classifying samples at different layers can boost the
performance of the model. Interestingly, we demonstrate that the CLDL
method is similar to constructing a conditional random field
(CRF)~\cite{crf} across multiple layers. In practice, the proposed
CLDL can be easily incorporated into most neural network architectures
trained using back propagation. We experimentally verify the
superiority of our method, achieving state-of-the-art performance
using various deep models, including NIN, GoogLeNet and VGGNet on six
heavily benchmarked datasets for object classification and scene
classification tasks.

The rest of this paper is organized as follows. Section \ref{sec:relatework}
reviews the related work. Detailed descriptions of CLDL
is given in Section~\ref{sec:method}. Experiments and discussions are presented in
Section~\ref{sec:experiments} and Section~\ref{sec:conclusion} concludes the
paper.

\vspace{-1mm}
\vspace{-3mm}
\section{Related Work}
\vspace{-2mm}
\label{sec:relatework}
Since Krizevsky et al.~\cite{krizhevsky2012imagenet} demonstrated the
dramatic performance improvement by deep networks in ImageNet
competition, deep networks have achieved exciting success in various
computer vision and machine learning tasks. Many factors are thought
to contribute to the success of deep learning, such as availability of
large-scale training datasets~\cite{imagenet,zhou2014learning}, deeper
and better network architectures~\cite{residual,vgg,googlenet},
development of fast and affordable parallel computing
devices~\cite{gpucomputing}, as well as a large number of effective
techniques in training large-scale deep networks, such as
ReLU~\cite{ReLU} and its variants~\cite{PReLU,SReLU},
dropout~\cite{dropout}, and data
augmentation~\cite{krizhevsky2012imagenet}. Here we mainly review
existing works that leverage multi-scale features learned at different
layers of a deep model and multiple objective functions to improve the
classification performance.

\vspace{1.5mm}
\noindent\textbf{Combining Multi-Scale Features} It is widely known that
different layers in a deep neural network output features with
different scales that represent the input data of various abstractness
levels. To boost the performance of deep networks, a natural idea is
to combine the complementary multi-scale features. Long et
al.~\cite{fullyconvseg} proposed to combine the features from multiple
layers and used the features to train a CRF for semantic
segmentation. Based on \cite{fullyconvseg}, Xie et al.~\cite{hed} used
multi-scale outputs of a deep network to perform edge
detection. Hypercolumn~\cite{hypercolumns} used the activations of CNN
units at the same location across all layers as features to boost
performance in segmentation and fine-grained localization. Similarly,
DAG-CNN~\cite{dagcnn} proposed to add prediction scores from multiple
layers as the final score in image classification. Different from the
above methods, our proposed CLDL method not only utilizes multi-scale
features by building classifiers on top of different layers, but also
encourages each classifier to automatically learn to specialize on
training patterns and concepts with certain abstractness during the
collaborative training. CLDL thus can effectively improve the overall
discriminative capability of the network.

\noindent\textbf{Combining Multiple Objective Functions} Some recent works propose to combine
multiple objective functions to train a deep model. In
\cite{multiloss}, several loss functions were appended to the output
layer of a deep network as regularizers to reduce its risk of
overfitting. DSN~\cite{dsn} proposed to add a ``companion'' hinge-loss
function for each hidden layer. Although the issue of ``vanishing
gradient'' in training can be alleviated, it is hard to evaluate the
contributions of the trained classifiers at hidden layers in DSN since
only the classifier at the output layer is used in
testing. GoogLeNet~\cite{googlenet} introduced classifiers at two
hidden layers to help speed up convergence when training a large-scale
deep network, and only used the classifier at the top output layer to
do inference. Different from the above methods, we propose a
collaborative objective function for multiple classifiers on different
layers, each of which coordinates with others to jointly train a deep
model and classify a new testing sample. A recent work of
LCNN~\cite{lcnn} aimed to improve the discriminability of the late
hidden layer by forcing each neuron to be activated for a manually
assigned class label. In contrast, our method has a stronger
discriminative capability by enabling each hidden layer to
automatically learn to be discriminative for certain data without
human interference.  \vspace{-3mm}
\begin{figure}[!t] \centering
    \vspace{-3mm}
	\includegraphics[width=11cm]{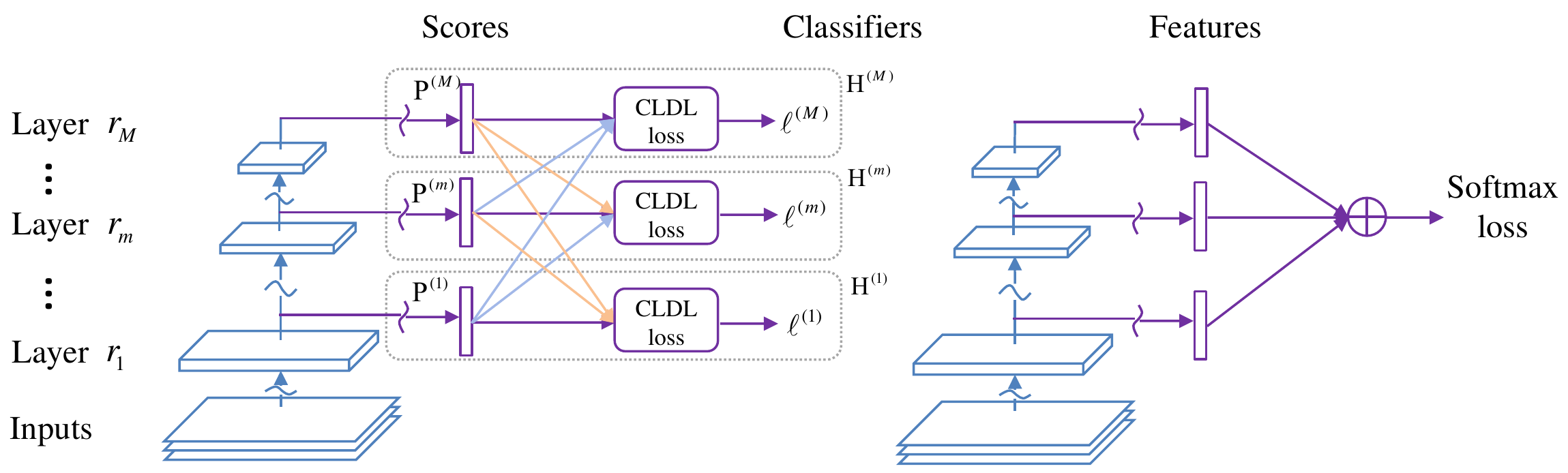}
	\vspace{-0.5mm}
	\caption{Comparisons on architectures of our proposed CLDL-DNN
          and related work \cite{dagcnn,hypercolumns}. The symbol
          ``$\sim$'' represents hidden layers. {\bf{Left}}:
          CLDL-DNN. Multiple classifiers
          ${\rm H}^{(1)}, \ldots, {\rm H}^{(M)}$ are introduced at
          different layers. With the CLDL loss
          (ref. Eqn. (\ref{eq:subeq})), each classifier is fed with
          the prediction scores from all other classifiers. We also
          introduce a simplified version of CLDL,
          i.e. CLDL$\raise1ex\hbox{-}$ in Section \ref{sec:cldls} by
          removing the feedback connections (orange lines). {\bf
            Right}: The architecture used in
          Hypercolumn~\cite{hypercolumns} and
          DAG-CNN~\cite{dagcnn}. Multi-scale features extracted at
          multiple layers are simply taken together and fed into a
          classifier, which has conventional loss functions,
          e.g. softmax loss. All notations in this figure are defined
          in the text of Section \ref{sec:formulation}.}
          \vspace{1mm}
	\label{fig:architecture}
	\vspace{-10mm}
\end{figure}

\noindent\section{Collaborative Layer-wise Discriminative Learning}
\label{sec:method}
\vspace{-2mm} In this section, we introduce our proposed CLDL method
in details. Firstly, we describe the motivation and definition of
CLDL. Secondly, we introduce the training and testing strategies of
deep models using CLDL. Thirdly, we explain the rationales for
CLDL. Fourthly, we give understanding on CLDL by establishing its
relation with classic conditional random fields
(CRF)~\cite{crf}. Finally, we explore variants of CLDL in order to
gain a deeper understanding of CLDL.  \vspace{-3.5mm}
\subsection{Motivation and Definition of CLDL}
\vspace{-1mm}
\label{sec:formulation}The proposed collaborative layer-wise
discriminative learning (CLDL) method aims to enhance the
discriminative capability of deep models by learning complementary
discriminative features at different layers such that each layer is
specialized for classifying samples of certain complexities. CLDL is
motivated by the widely recognized fact that the intermediate features
learned at different layers in a deep model are suitable for
discriminating visual patterns of different complexities. Therefore,
encouraging different layers to focus on categorizing input data of
different properties, rather than forcing each of them to address all
the data, one can improve layer-level discriminability as well as
final performance for a deep model. In other words, with this
strategy, the knowledge distributed in different layers of a deep
network can be effectively utilized and the discriminative capability
of the overall deep model is largely enhanced by taking advantage of
those discriminative features learned from multiple layers.

\vspace{0.8mm} We now give necessary notations to formally explain
CLDL. For brevity, we only consider the case of one training sample,
and the formulation for the multiple samples case can be derived
similarly since samples are independent. We denote a training sample
as $(\mathbf{x}, y^*)$ where $\mathbf{x}\in R^d$ denotes the raw input
data, $y^* \in \{1,\ldots,K\}$ is its ground truth category label and
$K$ is the number of categories.We consider a deep model consisting
of $L$ layers, each of which outputs a feature map denoted as
${\mathbf{X}}^{(l)}$. Here ${\mathbf{X}}^{(0)}$ and
${\mathbf{X}}^{(L)}$ represent the input and final output of the
network, respectively. $\mathbf{W}^{(l)}$ denotes the parameter of
filters or weights to be learned in the $l$-th layer. Using above
notations, the output of a $L$-layer deep network at each layer can
be written as \vspace{-0.5mm}
\begin{equation*} \label{activation}\footnotesize
\mathbf{X}^{(l)} = f^{(l)}(\mathbf{W}^{(l)}  * \mathbf{X}^{(l - 1)} ),\ \ l = 1, \ldots ,L\,
\ \ \text{and} \ \ \mathbf{X}^{(0)} \triangleq \mathbf{x},
\vspace{-2mm}
\end{equation*}
\vspace{-3.5mm}

\noindent where $f^{(l)}( \cdot )$ is a composite of multiple
specific functions including activation function, dropout,
pooling and softmax. For succinct notations, the bias
term is absorbed into $\mathbf{W}^{(l)}$.

\vspace{1.5mm} CLDL chooses $M$ layers out of the $L$ layers which are
indexed by $S = \{ r_m ,m = 1, \ldots ,M\}$,
$r_m \in \{1, \ldots, L\}$ and places classifiers on each of the
layers. Denote each classifier as ${\rm H}^{(m)}$ and the classifier
set excluding ${\rm H}^{(m)}$ as ${\rm\bar H}^{(m)}$. ${\rm H}^{(m)}$
outputs categorical probability scores
$\mathbf{P}^{(m)} = (\mathbf{P}^{(m)}(1), \ldots,
\mathbf{P}^{(m)}(K))$
over all $K$ categories. Note that we have
$\left\| {\mathbf{P}^{(m)} } \right\|_1 = 1$ since $\mathbf{P}^{(m)}$
denotes a probability distribution. When the classifier
${\rm H}^{(m)}$ has high confidence in classifying the input data
$\mathbf{X}^{(0)}$ to the category $y^*$, the value of
$\mathbf{P}^{(m)}(y^*)$ will be close to 1. The CLDL loss function for
${\rm H}^{(m)}$ is defined as \vspace{-3mm} {\footnotesize
  \setlength{\abovedisplayskip}{10pt}
  \setlength{\belowdisplayskip}{4pt}
\begin{align} \label{eq:subeq} \ell ^{(m)}(\mathbf{x}, y^*, \mathcal{W}) &= - \log \mathbf{P}^{(m)} (y^*
  )\prod\limits_{\scriptstyle t = 1, \scriptstyle t \ne m }^M
  {[1 - \mathbf{P}^{(t)} (y^* )]^{\frac{1}{{M - 1}}} },\\
\label{eq:tribeq}\footnotesize
\mathbf{P}^{(m)}(y^*)  &= h^{(m)}_{y^*} (\mathbf{w}^{(m)} ,\mathbf{X}^{(r_m )} ),
\end{align}
}%
\noindent where
$h^{(m)}_{y^*} (\mathbf{w}^{(m)} ,\mathbf{X}^{(r_m )} )$ denotes the mapping function of the classifier
${\rm H}^{(m)}$ from input feature $\mathbf{X}^{(r_m )}$ to
category label $y^*$, and $\mathbf{w}^{(m)}$ is the parameters associated with
${\rm H}^{(m)}$. $\mathcal{W}$ is defined as all the learnable weights in CLDL:

{ \setlength{\abovedisplayskip}{2.5pt}
  \setlength{\belowdisplayskip}{2.5pt}
\begin{equation*} \label{eq:allweights}\footnotesize {\mathcal{W}} = ({\bf{W}}^{(1)} ,
  \ldots ,{\bf{W}}^{(L)}, {\bf{w}}^{(1)}, \ldots, {\bf{w}}^{(M)} ).
\vspace{-2mm}
\end{equation*}
}
\vspace{-3mm}

For better understanding, we further divide the loss function in Eqn. (\ref{eq:subeq})
into multiplication of two terms as
\vspace{-2mm}

\begin{equation}
\label{eq:twoterm}\footnotesize
\ell ^{(m)}  = T^{(m)} C^{(m)},
\end{equation}
\vspace{-6.5mm}

\begin{equation}
\label{eq:Tm}\footnotesize
T^{(m)} = \prod\limits_{\scriptstyle
  t = 1, \scriptstyle t \ne m }^M {[1 - {\mathbf{P}}^{(t)} (y^*)]^{\frac{1}{{M - 1}}} }, \ \ \ \ \ \
 C^{(m)}  = - \log {\mathbf P^{(m)}} (y^*).
\vspace{-1mm}
\end{equation}

Here, $T^{(m)}$ carries modulation message collaborating with the
classifier ${\rm H}^{(m)}$, and $C^{(m)}$ is the confidence output by
${\rm H}^{(m)}$ (we discuss the roles of $T^{(m)}$ and $C^{(m)}$ in
more details later). Note that ${\rm H}^{(m)}$ employed in our method
can be chosen freely from many kinds of conventional classifiers to
satisfy the requirements of different tasks, including neural
network~\cite{booknn}, SVM~\cite{svm}, and logistic regression
classifier~\cite{logisticreg}, etc. The architecture of CLDL-DNN is
illustrated in Fig.~\ref{fig:architecture}.  \vspace{-3.5mm}

\subsection{Training and Testing Strategies for CLDL}
\vspace{-0.5mm}
The overall objective function of CLDL is a weighted sum
of loss functions from all classifiers, with a weight decay term
to control complexity of the model:
\vspace{-1.5mm}
\begin{equation*} \label{eq:overall} \footnotesize
  L^{(\text{Net})}(\mathbf{x},y^*,\mathcal{W}) = \sum\limits_{m = 1}^M {\lambda _m \ell
    ^{(m)} } + \mathbf{\alpha} \left\| {\mathcal{W} } \right\|_2,
\vspace{-1.5mm}
\end{equation*}

\noindent where $\mathbf{\alpha}\in R^+$ is the penalty factor which
is set to be the same for all learnable weights for simplicity, and
$\lambda_m \in R^+$ denotes the weight of each classifier, used to
balance the effect of the corresponding classifier in the overall
objective function. The goal of training is to optimize all the
learnable weights: \vspace{-0.8mm}
\begin{equation*}
\label{eq:opt}\footnotesize
\mathcal{W}^ *   = \mathop {\arg \min }\limits_\mathcal{W} L^{(\text{Net})} (\mathbf{x},y^*,\mathcal{W}).
\end{equation*}
\vspace{-3.1mm}

The network can be trained in an end-to-end manner by standard
back-propagation, and the gradient for variables of the $l$-th layer
\begin{scriptsize}
$Q^{(l)} \in \{ \mathbf{X}^{(l)} ,\mathbf{W}^{(l)} ,\mathbf{w}^{(l)} \}$
\end{scriptsize}
is calculated by following the chain rule which leads to
{\footnotesize
 \setlength{\abovedisplayskip}{5pt}
 \setlength{\belowdisplayskip}{0pt}
\begin{align}
\label{eq:gradients}
  \frac{{\partial L^{(\text{Net})} }}{{\partial
  Q^{(l)} }} &= \sum\limits_{m = 1}^M {\lambda _m \frac{{\partial \ell ^{(m)}
               }}{{\partial Q^{(l)} }}} + {\bf{\alpha}}\frac{{\partial \left\| {\mathcal{W} } \right\|_2 }}{{\partial
               Q^{(l)} }} = \sum\limits_{m = 1}^M {\lambda _m \frac{{\partial C ^{(m)} }}{{\partial Q^{(l)} }}} T^{(m)} + {\bf{\alpha}}\frac{{\partial \left\| {\mathcal{W} } \right\|_2 }}{{\partial Q^{(l)} }},\\[2pt]
\label{eq:gradientsx}
  \frac{{\partial \ell ^{(m)} }}{{\partial
  {\bf{X}}^{(l)} }} &=
    \begin{cases}
      \frac{-1}{{\mathbf{P}^{(m)} (y^*)}}\frac{{\partial h_{y^*}^{(m)} }}{{\partial \mathbf{X}^{(r_m )} }}\frac{{\partial f^{(r_m )} (\mathbf{W}^{(r_m)}  * \mathbf{X}^{(r_m - 1)} )}}{{\partial \mathbf{X}^{(l )} }}{T}^{(m)},& l < r_m  \\
      \frac{-1}{{\mathbf{P}^{(m)} (y^*)}}\frac{{\partial h_{y^*}^{(m)} }}{{\partial \mathbf{X}^{(r_m )} }} T^{(m)},& l = r_m  \\
      0, & l > r_m.  \\
   \end{cases}
\end{align}
}%

Recall $r_m$ is the index of the input layer for ${\rm H}^{(m)}$. The loss $\ell^{(m)}$
only contributes to optimizing the layers lying on the input layer of ${\rm
  H}^{(m)}$. Here,
{
\footnotesize
 \setlength{\abovedisplayskip}{5pt}
  \setlength{\belowdisplayskip}{4pt}
\begin{align}
\label{eq:gradientsW}
  \footnotesize
  \frac{{\partial \ell ^{(m)} }}{{\partial \mathbf{W}^{(l)} }} &= \frac{{\partial \ell^{(m)} }}{{\partial \mathbf{X}^{(r_m)} }}\frac{{\partial f^{(r_m)} (\mathbf{W}^{(r_m)}  * \mathbf{X}^{(r_m - 1)} )}}{{\partial \mathbf{W}^{(l)} }} + 2\alpha,l < r_m,\\[2pt]
\label{eq:gradientsw}\footnotesize
  \frac{{\partial \ell ^{(m)} }}{{\partial \mathbf{w}^{(l)} }} &= \frac{-1}{{\mathbf{P}^{(m)} (y^*)}}\frac{{\partial h_{y^*}^{(m)} }}{{\partial \mathbf{w}^{(l)} }} T^{(m)} + 2\alpha.
\end{align}
}%

In gradient calculation, we treat $T^{(m)}$ as independent of
$Q^{(l)}$ during the error back-propagation
w.r.t. $Q^{(l)}$. Therefore, we set
$\frac{{\partial T^{(m)} }}{{\partial Q^{(l)} }}=0$. In this way,
$T^{(m)}$ acts as a weight factor which is related with the prediction
scores output by classifiers in ${\rm \bar H}^{(m)}$ and it controls
the scale of the gradients calculated for updating ${\rm
  H}^{(m)}$.
The advantages of such simplification are two-fold. Firstly,
calculation of gradients becomes easy and fast, and meanwhile the
numerical problem in calculating
$\frac{{\partial T^{(m)} }}{{\partial Q^{(l)} }}$ when
$\mathbf{P}^{(s)}(y^*)$ for $s \in \{1,\ldots,M\}$ but $s \ne m$ is close to 1 can be avoided (see
Supplementary Materials for further details).  Secondly, it reduces
the risk of overfitting, which has been empirically verified and can be explained by seeing
$\frac{{\partial T^{(m)} }}{{\partial Q^{(l)} }}=0$ as a
regularizer. In practice, given the function forms of
${h_{y^*}^{(m)} }$ and $f^{(l)}$, it is easy to calculate necessary
gradients according to
Eqn. (\ref{eq:gradients})-(\ref{eq:gradientsw}).

In the training phase, we in fact optimize learn-able weights through a maximum likelihood estimation (MLE) as follows:

\begin{equation*}
  \label{eq:trainprob}\footnotesize
  {\mathcal W}^* = \mathop {\arg \max }\limits_{\mathcal W} P(y^*|\mathbf{x},{\mathcal{W}}) = \mathop {\arg \min }\limits_\mathcal{W}  L^{(\text{Net})}(\mathbf{x},y^*,\mathcal{W}),
\end{equation*}

\noindent where the likelihood distribution is parameterized by a deep
network. To be consistent with the training, in the testing phase, we
do inference to decide the most probable class label by solving the
discrete optimization problem

\begin{equation}
\label{eq:testprob}\footnotesize
y^* = \mathop {\arg \max }\limits_y P(y|\mathbf{x},\mathcal{W}) =　\mathop {\arg \min }\limits_y  L^{(\text{Net})}(\mathbf{x},y,\mathcal{W}),
\end{equation}

\noindent where $y \in \{1, \ldots, K\}$. Similarly, it is easy to
predict the top $k$ categories for the input data using
Eqn. (\ref{eq:testprob}).  \vspace{-3.2mm}

\subsection{Explanations on CLDL}
\vspace{-0.8mm}
In the following, we explain how the CLDL enhances the discriminative capability
of a deep network.

As shown in Eqn. (\ref{eq:twoterm}), the loss function of each
classifier considers two multiplicative terms, i.e.  $T^{(m)}$ and
$C^{(m)}$. Here, $C^{(m)}$, taking the form of entropy
loss~\cite{duda}, depicts the predicted confidence for the sample
belonging to a specific category. Minimizing $C^{(m)}$ pushes the
classifier ${\rm H}^{(m)}$ to hit its ground truth category. $T^{(m)}$
is a geometric mean of the prediction scores on the target class
output by other classifiers in ${\rm \bar H}^{(m)}$. $T^{(m)}$
measures how well those ``companion" classifiers perform on
classifying the input sample. Here comes the layer-wise collaboration
(or competition). When the input sample is correctly classified by all
classifiers in $\rm \bar H^{(m)}$, the value of $T^{(m)}$ is small;
otherwies, $T^{(m)}$ takes a large value. Considering $T^{(m)}$
together with $C^{(m)}$ distinguishes our CLDL loss function from
conventional loss functions: in CLDL, each classifier considers
performance of other classifiers in the same network when trying to
classify a input sample correctly. The classifier will put more
efforts on the samples difficult for other classifiers and care less
about samples that have been addressed well by other classifiers. As a
result, the optimization of CLDL can be deemed as a collaborative
learning process. All classifiers share a common goal: maximizing the
overall classification performance by paying attention to different
subsets of the samples. In more details, by using CLDL we encourage
the deep network to act in following ways.  \vspace{-0.5mm}
\begin{itemize}
\item If all classifiers in ${\rm \bar H}^{(m)}$ have correctly
  classified input data $\mathbf{x}$, we have $\mathbf{P}^{(t)} (y^*)$
  close to 1, for $t=1,\ldots,M$, but $t\ne m$. Hence $T^{(m)}$ takes
  a small value close to 0. According to
  Eqn. (\ref{eq:gradients})-(\ref{eq:gradientsw}), the gradients on
  learnable weights $\mathcal W$ that are back propagated from
  classifier ${\rm H}^{(m)}$ are suppressed by small $T^{(m)}$.  In
  other words, the classifier ${\rm H}^{(m)}$ at layer $r_m$ need not
  correctly predict $\mathbf{x}$ as it is informed that $\mathbf{x}$
  has already been classified correctly by other
  classifiers. Therefore, the risk of overfitting to these samples for
  ${\rm H}^{(m)}$ is reduced.  \vspace{0.5mm}
\item If no classifier in ${\rm \bar H}^{(m)}$ correctly predicts the
  category of the input $\mathbf{x}$, $T^{(m)}$ would have a large
  value close to 1. According to
  Eqn. (\ref{eq:gradients})-(\ref{eq:gradientsw}), ${\rm H}^{(m)}$
  will be encouraged to focus on learning this sample that is
  difficult for other classifiers. The hard sample may be well
  discriminated using the features of ${\rm H}^{(m)}$ at a proper
  level of feature abstraction.  \vspace{0.5mm}
\item If ${\rm \bar H}^{(m)}$ is a mixture of classifiers, some of which correctly
  classify the input and some cannot. Then one can see that the value of
  $T^{(m)}$ is positively correlated with the prediction score of
  ${\rm H}^{(m)}$ on the ground truth category (see Supplementary Materials for
  rigorous derivations). Thus the classifier with the highest prediction score
  will dominate the updating of the weights. In this way, we encourage the
  classifier with the best discriminative capability to play the most important
  role in learning from the input data.
\end{itemize}
\vspace{-0.5mm}

We also note that other methods that add conventional classifiers on
multiple layers, e.g. GoogLeNet~\cite{googlenet} and DSN~\cite{dsn}
can be viewed as special cases of our method by setting $T^{(m)}$ as a
constant 1 and the values of $\lambda_m$ for classifiers at hidden
layers as 0 in testing. In \cite{googlenet,dsn}, since no classifier
stays informed of the output of other classifiers, every classifier is
forced to fit all of the training data and ignores the different
layer-wise discriminative capabilities to different input data. One
disadvantage of such a strategy is that classifiers would be prone to
overfitting, thus hampering the discriminability of the overall
model. In contrast, by focusing on learning from certain samples,
classifiers in CLDL reduce the risk of overfitting over the whole
training set and have a better chance to learn more discriminative
representations for the data.  \vspace{-3mm}
\subsection{Discussions on Relation Between CLDL and CRF}
\vspace{-0mm} In this subsection, we demonstrate that CLDL can be
viewed as a simplified version (with higher optimization efficiency)
of a conditional random field (CRF) model.

CRF is an undirected graphical discriminative model that compactly
represents the conditional probability of a label set
$Y=\{y^*_1,\ldots,y^*_n\}$ given a set of observations
$X = \{\mathbf{x}_1,\ldots,\mathbf{x}_n\}$, i.e. $ P(Y|X) $. In CLDL,
we introduce another hidden label set $S=\{s_1,\ldots,s_n\}$ to be the
assignment of each $\mathbf{x}_i \in X$ to a certain classifier
${\rm H}^{(m)}$. $s_i$ takes its value from $\{1,\ldots,M\}$. Recall
$M$ is the number of classifiers in CLDL. In our classification
scenario, given a training set
$ \{(\mathbf{x}_1,y^*_1),\ldots,(\mathbf{x}_n,y^*_n)\} $, optimizing
$ P(Y|X) = \sum_{S} P(Y|X,S)P(S|X) $ w.r.t the weight parameter gives
a CRF model that distributes $n$ observations into $M$ classifiers
with an optimal configuration in the sense of maximizing the training
accuracy.

 More concretely, the conditional probability specified in our CRF model can be written as
\begin{equation*}
 \footnotesize
 P(S|X) = \frac{1}{Z} \exp(\beta^\top f(S,X) ),
 \end{equation*}
  where $Z$ is a partition function and $\beta$ is the weight parameter.
 Following the notations given in Eqn.~\eqref{eq:tribeq}, the function $f(S,X)$ in our CLDL case is specifically defined as
\begin{equation*}
\footnotesize
f(S,X) = \left[\log \left(1-h_{y_i}^{1}({\bf w}^{(1)},\mathbf{X}^{(r_1)})\right), \ldots, \log \left(1-h_{y_i}^{M}({\bf w}^{(M)},\mathbf{X}^{(r_M)})\right)\right]^\top,
\vspace{-2mm}
\end{equation*}
and each element of the weight parameter $\beta$ takes a fixed value $\frac{1}{M-1}$.

Then the likelihood $ P(Y|X)$ is given by classifiers associated with the layers indicated by $S$. Here, $P(Y|X,S)$ is parameterized by the chosen classifier as indicated in Eqn.~\eqref{eq:tribeq}:
$P(y^*|\mathbf{x},s) = h^{(s)}_{y^*}({\mathbf w}^{(s)},\mathbf{X}^{(r_s)})$.
Maximizing $ P(Y|X)$ gives the optimal value of the assignment indicator $s$ for $\bf x$ as well as the classifier parameter ${\bf w}^{(s)}$ for each collaborative classifier.

CRF can be solved via a standard message passing algorithm. In CLDL, we simplify the CRF into a chain and apply error back propagation for optimization.
\vspace{-8.2mm}
\subsection{Variants of CLDL}
\label{sec:cldls}
\vspace{-1.8mm}
To further verify the effectiveness of CLDL, we have also explored an
alternative method to utilize the layer-level discriminative information and we
here compare it with CLDL.

This method we explore is called CLDL$\raise1ex\hbox{-}$ and can be seen as a simplification of CLDL. As indicated in
Fig.~\ref{fig:architecture}, its only difference from CLDL lies in that there is no feedback connection from classifiers at top layers to classifiers at bottom layers. More concretely, in the
definition of $T^{(m)}$ for CLDL$\raise1ex\hbox{-}$, which is formulated by
$T^{(m)} = \prod\limits_{t = 1}^{m - 1} {(1 - {\bf P}^{(t)} (y^*))^{\frac{1}{{m
        - 1}}} } $,
we can see that the information flow among different classifiers takes a
single direction: the classifiers on top layers can get the prediction scores
from classifiers on bottom layers, but the reverse does not hold. This is similar
to the cascading strategy used in face detection~\cite{violajones}. The advantage of CLDL over CLDL$\raise1ex\hbox{-}$ is that each
classifier is able to automatically focus on learning to categorize certain examples by taking all other
classifiers' behavior into optimization. Therefore, CLDL demonstrates better
discriminative capability than CLDL$\raise1ex\hbox{-}$, which is empirically verified in the experiment part.
\vspace{-3mm}
\section{Experiments and Analysis}
\vspace{-2mm}
\label{sec:experiments}
\subsection{Experimental Setting} \vspace{-1mm}To evaluate our method thoroughly, we conduct
extensive experiments for two classification tasks, i.e. object
classification on CIFAR-100~\cite{cifar}, MNIST
\noindent\cite{MNIST} and ImageNet~\cite{imagenet} datasets, and scene
classification on MIT67~\cite{mit67}, SUN397~\cite{sun397} and
Places205~\cite{zhou2014learning} datasets. There are overall three
state-of-the-art deep neural networks with different architectures tested on
these datasets, including NIN~\cite{NIN}, GoogLeNet~\cite{googlenet} and
VGGNet~\cite{vgg}. Specifically, NIN is used on CIFAR-100 and MNIST, GoogLeNet
is used on ImageNet and VGGNet is used on scene recognition tasks. All of these deep models have achieved state-of-the-art
performance on the datasets we use. We choose Caffe~\cite{caffe} as the platform to train different models and
conduct our experiments. To reduce the training time, four NVIDIA TITAN X GPUs
are employed in parallel for training.

\begin{figure}[h]
\vspace{-2mm}
\centering
\begin{subfigure}{0.45\textwidth}
\centering
\includegraphics[scale=0.4]{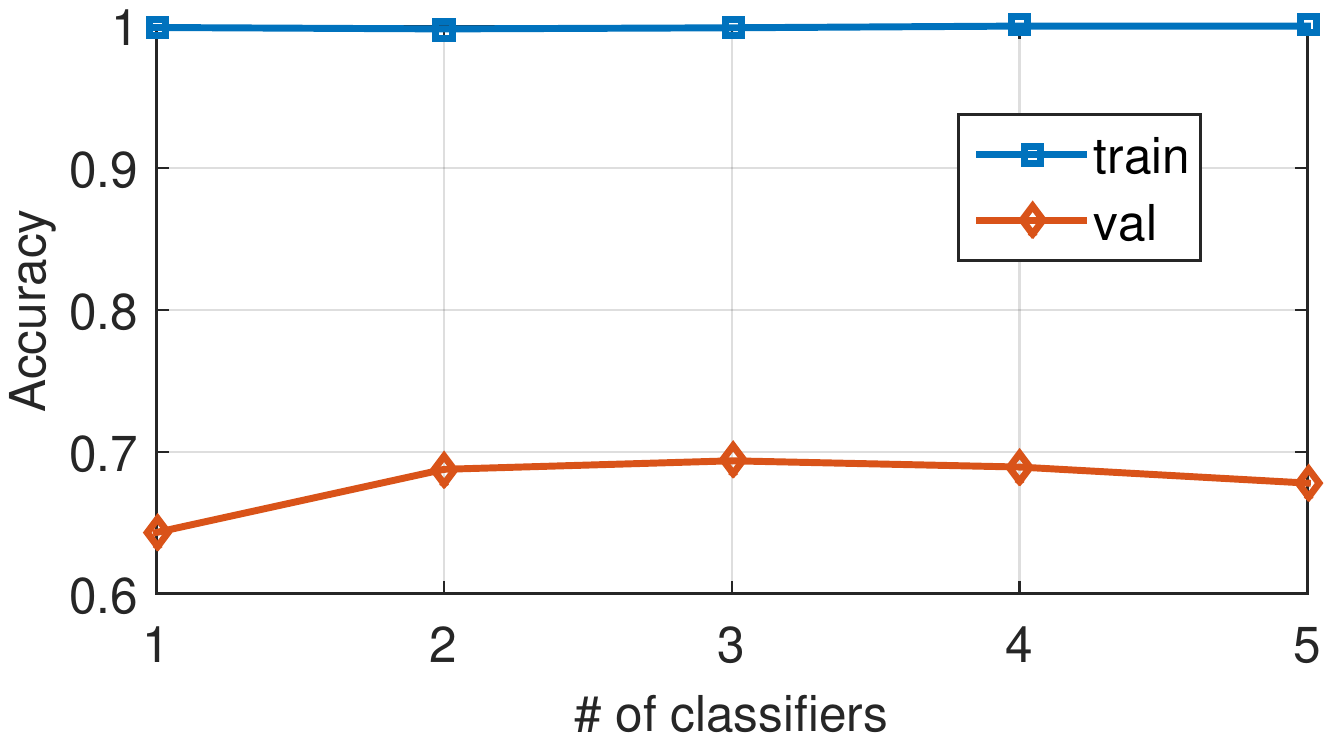}
\end{subfigure}
\begin{subfigure}{0.45\textwidth}
\centering
\includegraphics[scale=0.4]{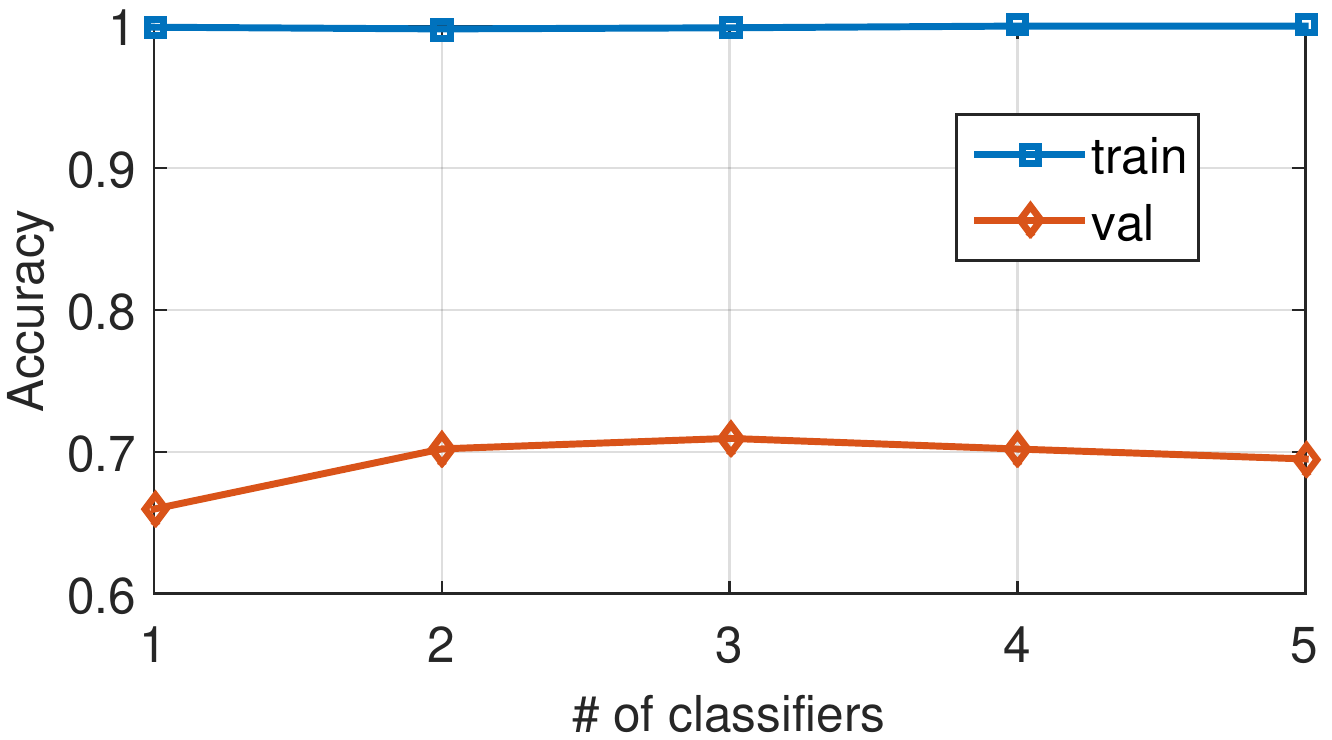}
\end{subfigure}
\vspace{-1mm}

\caption{Evaluations of NIN model on CIFAR-100 for investigating the effect of
  classifier number in CLDL on the classification accuracy for both training
  and validation sets. \textbf{Left}: classification accuracy curves without data
  augmentation. \textbf{Right}: classification accuracy curves with data augmentation.}
\label{fig:eval_clsnum}
\vspace{-6mm}
\end{figure}
\vspace{-4mm}
\subsection{Deciding Position and Number of Classifiers}
Before applying CLDL in practice, an important problem one needs to solve is
to determine which layers to put the collaborative classifiers. Analytically
solving this problem is hard. Therefore, we propose a simple yet effective
heuristic method to determine the proper position and number of
classifiers. From top output layer to bottom input layer, we place a classifier
every $V$ weight layers and $V$ is calculated by $V = \left\lceil {\left( L/M \right)^\gamma  } \right\rceil$, in which $L$ and $M$ follow the notations in Section \ref{sec:formulation}. Accordingly, indexes of layers to put classifiers are calculated by $r_m = L-(M-m)V$, $m \in {1,\ldots,M}$. Throughout our experiments, we set $\gamma < 1$ (here $\gamma = 0.8$) to suppress the
value of $V$. In this way, one can avoid placing classifiers at very bottom
layers when the number of classifiers is large, because very bottom layers
describe basic concepts and should be shared among all categories. To test the
influence of various numbers of collaborative classifiers in CLDL on the final performance, we
conduct primitive experiments on CIFAR100 with data augmentation using different
numbers of classifiers ($M=1,2,3,4,5$). When $M=1$, we are actually using
a single softmax classifier on top of the network. The experimental results are shown in
Fig. \ref{fig:eval_clsnum}, from which one can observe that the performance
increases with more classifiers added (when $M=1,2,3$) at the beginning, and
then decreases if we continue adding more classifiers (when $M=4,5$). This
phenomenon could be explained as follows: when the number of classifiers is
small, the deep network can benefit from various discriminative information in
different layers, but the network will gain little from too many
classifiers added because neighboring layers often contain redundant information with
each other. Finally, the performance on the validation set will drop due to
overfitting. Based on the conclusion from this experiment, in the following
experiments, we set $M=3$ for all the datasets. Note that more careful tuning on the
number of classifiers for different datasets might further improve the
performance of CLDL. Nevertheless, we show that state-of-the-art performance
has been achieved using the same configuration.

In our experiments, three kinds of deep models including NIN, GoogLeNet and
VGGNet are used as base models to evaluate the performance of CLDL on
different datasets. We denote those three models trained with CLDL as CLDL-NIN,
CLDL-GoogLeNet and CLDL-VGGNet, respectively. The positions of classifiers are in line with the calculation of $r_m$. In CLDL-NIN, each classifier
consists of a mlpconv layer~\cite{NIN} to output feature maps with the same
number of channels as the number of categories, and global averaging pooling
layer~\cite{NIN} to transform the size of the feature map into $1 \times
1$. In
CLDL-GoogLeNet, we just simply replace the softmax
loss function in each classifier with CLDL loss function given in
Eqn. (\ref{eq:subeq}) without changing the rest network structure. By doing
so, the results of CLDL-GoogLeNet can show more clearly the effects of CLDL on a
deep model. In CLDL-VGGNet, similar to previous methods such as
\cite{dagcnn,hypercolumns}, we use two fully
connected layers in our classifiers. Throughout experiments, we set
$\lambda_1=\lambda_2=0.3, \lambda_3=1$. 

\vspace{-2mm}
\begin{table}[!t]
\setlength{\tabcolsep}{10pt}
\footnotesize
\caption{Classification error rates on CIFAR-100 either when data augmentation is used or not. d.a. represents ``data augmentation". Note since there is no reported result on CIFAR-100 using data augmentation in NIN~\cite{NIN}, we refer to the results reimplemented by \cite{apl} (denoted by NIN$^*$). All tespts are by single model and single~crop.} \centering
\vspace{2mm}
	\label{table:cifar100}
	\begin{tabular}{lcc} \hline Model  & Without d.a. (\%) & With d.a. (\%)\\
          \hline  Maxout~\cite{maxout}              & 38.57        & -\\
          Prob maxout~\cite{probmaxout}             & 38.14        & - \\
          Tree based priors~\cite{treebased}        & 36.85        & - \\
          CNN + Maxout~\cite{dasnet}                & -            & 34.54 \\
          dasNet~\cite{dasnet}                      & -            & 34.50 \\
          NIN~\cite{NIN}                            & 35.68        & - \\
          NIN$^*$~\cite{apl}                        & 35.96        & 32.75 \\
          APL~\cite{apl}                            & 34.40        & 30.83 \\
          DSN~\cite{dsn}                            & 34.57        & - \\
          \hline
          $\rm{DSN}^{*}$-NIN (ours)                  & 34.12        & 32.95 \\
          CLDL$\raise1ex\hbox{-}$-NIN	(ours)              & 31.27        & 30.41 \\
          CLDL-NIN (ours)			    & \textbf{30.40} & \textbf{29.05} \\
          \hline
	\end{tabular}
	\label{table:CIFARRESULT}
	\vspace{-6mm}
\end{table}

\begin{table}

\begin{minipage}[l]{0.49\textwidth}%
\footnotesize
  \caption{Classification error rates on MNIST. All tests are by single model
    and single crop.}
    \vspace{1mm}
	\label{table:mnist}
	\centering
	\begin{tabular}{lc} \hline Model     & {Error rate (\%)} \\
		\hline
		Stochastic Pooling~\cite{stochasticpooling}     &  0.47 \\
		Maxout~\cite{maxout}                            &  0.47 \\
		NIN~\cite{NIN}                                  &  0.42 \\
		DSN~\cite{dsn}                                  &  0.39 \\
		CLDL-NIN (ours)                                 &  \textbf{0.28} \\
		\hline
	\end{tabular}
\end{minipage}\ \ \ \ \ \
\begin{minipage}[r]{0.49\textwidth}%
\footnotesize
  \caption{Top-5 classification error rates on ImageNet. Tests for CLDL-GoogLeNet are
    by single model and single crop.}
    \vspace{1mm}
	\label{table:imagenet}
	\centering
	\begin{tabular}{lc} \hline Model & Top-5 (\%)\\ \hline
		AlexNet~\cite{krizhevsky2012imagenet}	      & 15.4      \\
		ZF~\cite{ILSVRC15}                            & 13.51     \\
		LCNN~\cite{lcnn}                              & 12.91     \\
		GoogLeNet\footnotemark~\cite{googlenet}       & 11.1      \\
		CLDL-GoogLeNet (ours)                         & \textbf{10.21} \\
		\hline
	\end{tabular}
\end{minipage}
\vspace{-4mm}
\end{table}
\footnotetext{$\textrm{https://}\textrm{github}.\textrm{com/}\textrm{BVLC/}\textrm{caffe/}\textrm{tree/}\textrm{master/}\textrm{models/}\textrm{bvlc}_{\textrm{--}}\textrm{googlenet}$}

\subsection{Results for Object Classification}

We now apply CLDL to object recognition on the following three benchmark datasets. All of our
models using CLDL are trained from scratch.
\vspace{-5mm}
\subsubsection{CIFAR-100}
The CIFAR-100 dataset contains 50,000 and 10,000 color images with size of
$32\times32$ from 100 classes for training and testing purposes,
respectively. Following~\cite{maxout}, preprocessing 
including global contrast normalization and ZCA whitening is applied. The
comparison results of CLDL-NIN with other state-of-the-art models with and without
data augmentation are shown in Table \ref{table:CIFARRESULT}, from which we can see that CLDL-NIN
achieves the best performance against all the compared methods.

Specifically, CLDL remarkably outperforms the baseline model (NIN) by reducing
the error rates by 5.56\%/3.70\% with/without data augmentation,
demonstrating the effectiveness of CLDL in enhancing the
discriminative capability of deep models. Compared with DSN, which imposes
independent classifiers on each hidden layer of NIN, CLDL-NIN reduces the error
rate by 4.17\% when no data augmentation is used. Furthermore, we replace the
loss function of each classifier in CLDL-NIN by conventional softmax loss
function, which gives $\rm{DSN}^{*}$-NIN. We train $\rm{DSN}^{*}$-NIN using the
training methods for DSN. By comparing the performance of CLDL-NIN and
$\rm{DSN}^{*}$-NIN, we can see CLDL-NIN achieves lower error
rates either when data augmentation is used or not. This clearly proves that our
method has superiority on improving the discriminative capability of the deep
model and alleviating overfitting (both models achieve nearly 100\% accuracy on
the training set) through allowing the classifiers to work collaboratively. Besides, compared with CLDL$\raise1ex\hbox{-}$, CLDL-NIN further reduces the error rates by 0.87\%/1.36\% with/without data augmentation, proving the advantages of CLDL over CLDL$\raise1ex\hbox{-}$.
\vspace{-5mm}
\subsubsection{MNIST}
MNIST is a heavily benchmarked dataset, which contains 70,000 28$\times$28 gray scale
images of numerical digits from 0 to 9, splitting into 60,000 images for
training and 10,000 images for testing.  On this dataset, we  apply neither any
preprocessing to the image data nor any data augmentation method, both of
which may further improve the performance. A summary of best methods on this
dataset is provided in Table \ref{table:mnist}, from which one can again observe that
CLDL-NIN performs better than other methods with a significant margin.
\vspace{-5mm}
\subsubsection{ImageNet}
To test the scalability of our method to a large number of classes and deeper
networks, we evaluate the CLDL method with a much more challenging and
larger-scale 1000-class ImageNet dataset, which contains roughly 1.2 million
training images, 50,000 validation images and 100,000 test images. Our baseline
model is the GoogLeNet, which has reported the best performance on image
classification in the ImageNet competition in 2014~\cite{ILSVRC15}. We train CLDL-GoogLeNet from
scratch using the publicly available configurations released by Caffe in Github$^1$. On
this dataset, no additional preprocessing is used except subtracting the
image mean from each input raw image.

Table \ref{table:imagenet} summarizes the performance of CLDL-GoogLeNet and
other deep models on the validation set of ImageNet. Compared with the original
GoogLeNet model$^1$ released by Caffe, CLDL-GoogLeNet achieves a 0.89 point
boost on this challenging dataset. Particularly, our method significantly
surpasses recently proposed LCNN~\cite{lcnn} which adds explicit supervision to
hidden layers of GoogLeNet. Some examples corretly classified by CLDL-GoogLeNet are illustrated in Fig.~\ref{fig:featurevis}. \vspace{-7.5mm}
\subsection{Results for Scene Classification}
\vspace{-0.5mm}
\begin{table}[!t]
\setlength{\tabcolsep}{6pt}
\footnotesize
\caption{Classification error rates on SUN397, MIT67 and Places205 datasets. For the former two datasets, top-1 accuracy rates are reported, while for the last dataset, we report the top-5 error rates. For models: VGGNet$_{ft}$-(11/16/11) and CLDL-VGGNet-(11/16/11), the three numbers separated by slash in brackets represent the sizes of the VGGNets that are used in training corresponding datasets. Please see text for details.}
\vspace{1mm}
\label{table:sceneresults}
\centering
\begin{tabular}{lccc}
  \hline
  Model  & MIT67 (\%)  & SUN397 (\%)  & Places205 (\%)\\
  \hline
  Places~\cite{zhou2014learning}                    & 54.32        & 68.24        & 50.00  \\
  Caffe~\cite{dagcnn}                               & 59.50        & 43.50        & -      \\
  Deep19~\cite{dagcnn}                              & 70.80        & 51.90        & -      \\
  Places205-AlexNet~\cite{zhou2014learning}         & 68.20        & 54.30        & 80.90  \\
  Places205-GoogLeNet~\cite{leaderboard}            & 76.30        & 61.10        & 85.41  \\
  Places205-CNDS-8~\cite{cnds}                      & 76.10        & 60.70        & 84.10  \\
  DAG-CNN~\cite{dagcnn}                             & 77.50        & 56.20        & -      \\
  Places205-VGGNet-11~\cite{wangplace205}           & 82.00        & 65.30        & 87.60  \\
  Places205-VGGNet-13~\cite{wangplace205}           & 81.90        & 66.70        & 88.10  \\
  Places205-VGGNet-16~\cite{wangplace205}           & 81.20        & 66.90        & 88.50  \\
  \hline
  VGGNet$_{ft}$-(11/16/11)                     & 83.10        & 68.47        & 87.60  \\
  CLDL-VGGNet-(11/16/11)(ours)                      & \textbf{84.69} &\textbf{70.40} & \textbf{88.67}  \\
  \hline
\end{tabular}
\vspace{-5.4mm}
\end{table}

Compared with object-centric classification tasks, scene classification is more
challenging because scene categories have larger degrees of
freedom. Recognizing different scenes needs the understanding  of the containing objects
(object-level) as well as their spatial relationships
(context-level). Therefore, to achieve good performance on this task, deep
networks are required to have strong discriminative capability on different levels
of representations.

In the following experiments, 
we take advantage of the publicly available pre-trained
Places205-VGGNet\footnote{$\textrm{https://}\textrm{github}.\textrm{com/}\textrm{wanglimin/}\textrm{Places205-VGGNet}$}
models in \cite{wangplace205} to verify effectiveness of CLDL on various scene
classification datasets. We use the strategy of fine-tuning to train deep models after using the
collaborative classifiers. Specifically, among all Places205-VGGNet models with
different depths ($\#$ of layers: 11, 13 and 16), Places205-VGGNet-11
and Places205-VGGNet-16 models are used as base models in our method as they
have achieved the best results on the MIT67 and SUN397 datasets accroding
to~\cite{wangplace205}, respectively. Since Places205 is a large-scale dataset
and it is time-consuming to train deep models from scratch, we fine-tune the
Places205-VGGNet-11 model using CLDL and achieve even better results than deeper
models, e.g. Places205-VGGNet-13 and Places205-VGGNet-16. For fair comparison,
we also fine-tune the models$^2$ on all tested datasets and compare their results
with ours. The fine-tuned models are denoted as Places205-VGGNet$_{ft}$. Similar
to~\cite{wangplace205}, we follow the multi-view classification method by averaging
the 10 prediction values from four corners and center of the image and their
horizontally flipped version.  \vspace{-5mm}
\subsubsection{SUN397}
SUN397 is a large scene recognition dataset with 130K images spanning 397
categories.  Seen from table
\ref{table:sceneresults}, CLDL-VGGNet-11 achieves the best performance among all
compared methods. Particularly, compared with DAG-CNN~\cite{dagcnn}, which
combines the multi-scale features from multiple hidden layers in VGGNet-19 to
perform classification, our method surpasses it significantly (14.2\%) with less weight layers, which verifies the
effectiveness of enhancing the discriminative capability of a deep model using our method.
\vspace{-5mm}
\subsubsection{MIT67}
MIT67 contains 67 indoor categories, with 15k color images. The standard training/testing datasets consist
of 5,360/1,340 images. Again, our CLDL-VGGNet-16 achieves the best result vs
other methods, establishing a new state-of-the-art for this challenging dataset.
\vspace{-5mm}

\subsubsection{Places205}
We also verify our method on Places 205, which is a much more challenging scene
recognition dataset compared with MIT67 and SUN397. It contains over 2.4
million images from 205 scene categories as the training set and 20,500 images as the
validation set. By comparison, our CLDL-VGGNet-11 not only outperforms the
original Places205-VGGNet-11 model by 1.07\%, but also achieves even better
performance compared to deeper networks, i.e. Places205-VGGNet-13,
Places205-VGGNet-16, which demonstrate that our methods can effectively improve the
performance of state-of-the-art deep models.
\vspace{-3mm}
\section{Conclusion and Future Work}
\label{sec:conclusion}
\vspace{-1mm}
In this paper, we propose a novel learning method called {\bf C}ollaborative {\bf L}ayer-wise {\bf D}iscriminative {\bf L}earning (CLDL) to enhance the
discriminative capability of a deep model. Multiple collaborative classifiers
are introduced at multiple layers of a deep model. Using a novel CLDL-loss
function, each classifier takes input not only the features from its input layer in the network, but also the prediction scores from other companion
classifiers. All classifiers coordinate with each other to jointly maximize the
overall classification performance. In future work, we plan to apply our method
to other machine learning tasks, e.g. image captioning.
\newpage


\section*{Supplementary material}

# Supplementary Material

**Abstract** In this supplementary material, we provide proofs for following two propositions proposed in Section 3.2 and Section 3.3.

**Proposition 1.** There is a numerical problem in calculating $\frac{\partial T^{(m)}}{\partial Q^{(l)}}$ when $\mathbf{P}^{(s)}(y^*)$ is close to 1 for $s \in \{1, \ldots, M\}$ but $s \neq m$, which can be avoided by setting $\frac{\partial T^{(m)}}{\partial Q^{(l)}} = 0$ in CLDL (Section 3.2).

*Proof.* Recall that the overall objective function of CLDL is

$$L^{(\text{Net})}(\mathbf{x}, y^*, \mathcal{W}) = \sum_{m=1}^{M} \lambda_m \ell^{(m)} + \alpha \|\mathcal{W}\|_2,$$

and the loss function for each collaborative classifier is

$$\ell^{(m)} = T^{(m)} C^{(m)},$$

where

$$T^{(m)} = \prod_{t=1, t \neq m}^{M} [1 - \mathbf{P}^{(t)}(y^*)]^{\frac{1}{M-1}}, \qquad C^{(m)} = -\log \mathbf{P}^{(m)}(y^*). \tag{1}$$

By taking account of $\frac{\partial T^{(m)}}{\partial Q^{(l)}}$, the gradient of the loss for variable $Q^{(l)} \in \{\mathbf{X}^{(l)}, \mathbf{W}^{(l)}, \mathbf{w}^{(l)}\}$ is calculated by following the chain rule:

$$\frac{\partial L^{(\text{Net})}}{\partial Q^{(l)}} = \sum_{m=1}^{M} \lambda_m \left( \frac{\partial C^{(m)}}{\partial Q^{(l)}} T^{(m)} + \frac{\partial T^{(m)}}{\partial Q^{(l)}} C^{(m)} \right) + \alpha \frac{\partial \|\mathcal{W}\|_2}{\partial Q^{(l)}},$$

where

$$\begin{aligned}
\frac{\partial T^{(m)}}{\partial Q^{(l)}} &= \frac{1}{M-1} \frac{\prod_{t=1, t\neq m}^{M}[1-\mathbf{P}^{(t)}(y^*)]^{\frac{1}{M-1}}}{\prod_{t=1, t\neq m}^{M}[1-\mathbf{P}^{(t)}(y^*)]} \frac{\partial \prod_{t=1, t\neq m}^{M}[1-\mathbf{P}^{(t)}(y^*)]}{\partial Q^{(l)}} \\
&= \frac{1}{M-1} \frac{\prod_{t=1, t\neq m}^{M}[1-\mathbf{P}^{(t)}(y^*)]^{\frac{1}{M-1}}}{\prod_{t=1, t\neq m}^{M}[1-\mathbf{P}^{(t)}(y^*)]} \sum_{s=1, s\neq m}^{M} \frac{\partial[1-\mathbf{P}^{(s)}(y^*)]}{\partial Q^{(l)}} \prod_{t=1, t\neq s, t\neq m}^{M}[1-\mathbf{P}^{(t)}(y^*)] \\
&= \frac{1}{M-1} \sum_{s=1, s\neq m}^{M} \frac{\partial[1-\mathbf{P}^{(s)}(y^*)]}{\partial Q^{(l)}} \frac{\prod_{t=1, t\neq s, t\neq m}^{M}[1-\mathbf{P}^{(t)}(y^*)] \prod_{t=1, t\neq m}^{M}[1-\mathbf{P}^{(t)}(y^*)]^{\frac{1}{M-1}}}{\prod_{t=1, t\neq m}^{M}[1-\mathbf{P}^{(t)}(y^*)]} \\
&= \frac{1}{M-1} \sum_{s=1, s\neq m}^{M} \frac{\partial[1-\mathbf{P}^{(s)}(y^*)]}{\partial Q^{(l)}} \frac{\prod_{t=1, t\neq m}^{M}[1-\mathbf{P}^{(t)}(y^*)]^{\frac{1}{M-1}}}{1-\mathbf{P}^{(s)}(y^*)} \\
&= \frac{1}{M-1} \sum_{s=1, s\neq m}^{M} \frac{\partial[1-\mathbf{P}^{(s)}(y^*)]}{\partial Q^{(l)}} \frac{\prod_{t=1, t\neq s, t\neq m}^{M}[1-\mathbf{P}^{(t)}(y^*)]^{\frac{1}{M-1}}}{[1-\mathbf{P}^{(s)}(y^*)]^{\frac{M-2}{M-1}}}.
\end{aligned} \tag{2}$$

From Eqn. (2), we can conclude that when $\mathbf{P}^{(s)}(y^*)$ is close to 1 for $s \in \{1, \ldots, M\}$ but $s \neq m$, the denominator $[1 - \mathbf{P}^{(s)}(y^*)]^{\frac{M-2}{M-1}}$ will be close to 0 in our experiments ($M \geq 3$), which incurs a numerical problem during the calculation of gradients. By setting $\frac{\partial T^{(m)}}{\partial Q^{(l)}} = 0$, such a numerical problem can be avoided. Meanwhile, as verified in our experiments, setting $\frac{\partial T^{(m)}}{\partial Q^{(l)}} = 0$ does not negatively affect the optimization of deep models since each classifier can still collaborate with other classifiers via the modulation message carried by $T^{(m)}$. Note that when $M = 1$, i.e. there is only one classifier in CLDL, the loss function for the classifier is conventional softmax loss function. When $M = 2$, although the numerical problem does not exist because the denominator $[1 - \mathbf{P}^{(s)}(y^*)]^{\frac{M-2}{M-1}}$ is equal to 1, the performance is inferior compared with the case when $M \geq 3$ according to the evaluation results displayed in Fig. 3 in the original text. □

**Proposition 2.** The value of $T^{(m)}$ is positively correlated with the prediction score of $\mathrm{H}^{(m)}$ on the ground truth category, i.e. $\mathbf{P}^{(m)}(y^*)$ (Section 3.3).

*Proof.* According to the definition of $T^{(m)}$ in Eqn. (1), we have

$$\begin{aligned} T^{(m)} &= \prod_{t=1,t\neq m}^{M} [1 - \mathbf{P}^{(t)}(y^*)]^{\frac{1}{M-1}} \\ &= \frac{\prod_{t=1}^{M} [1-\mathbf{P}^{(t)}(y^*)]^{\frac{1}{M-1}}}{[1-\mathbf{P}^{(m)}(y^*)]^{\frac{1}{M-1}}} \\ &= \left\{ \frac{\prod_{t=1}^{M} [1-\mathbf{P}^{(t)}(y^*)]}{1-\mathbf{P}^{(m)}(y^*)} \right\}^{\frac{1}{M-1}} \\ &= \left[ \frac{\mathcal{G}}{1-\mathbf{P}^{(m)}(y^*)} \right]^{\frac{1}{M-1}}, \end{aligned} \qquad (3)$$

where $\mathcal{G} = \prod_{t=1}^{M} [1 - \mathbf{P}^{(t)}(y^*)]$ is a constant for each collaborative classifier $\mathrm{H}^{(m)}$. As observed from Eqn. (3), the value of $T^{(m)}$ is monotonically increasing (positively correlated) with $\mathbf{P}^{(m)}(y^*)$. □


\clearpage
\bibliographystyle{splncs}
\bibliography{mybib}
\end{document}